\documentclass{article}

\PassOptionsToPackage{round}{natbib}

\usepackage[preprint]{neurips_2026}

\usepackage[utf8]{inputenc} 
\usepackage[T1]{fontenc}    
\usepackage{hyperref}       
\usepackage{url}            
\usepackage{booktabs}       
\usepackage{amsfonts}       
\usepackage{nicefrac}       
\usepackage{microtype}      
\usepackage{xcolor}         
\usepackage{amsthm}
\usepackage{graphicx}
\usepackage{amsmath}
\usepackage{algorithm}
\usepackage{algpseudocode}
\usepackage{wrapfig}
\usepackage{placeins}
\usepackage{appendixautoref}

\theoremstyle{definition}

\hypersetup{
    colorlinks,
    linkcolor={blue!55!black},
    citecolor={gray!55!black},
    urlcolor={blue!80!black}
}

\title{Learning to Solve Hard Problems in RL for LLMs \\ by Never Giving Up}

\author{%
  \parbox[t]{0.36\textwidth}{\centering
    Michael Noukhovitch\thanks{Correspondence to \texttt{mnoukhov@gmail.com},
    see code at \href{https://github.com/mnoukhov/never-give-up}
    {github.com/mnoukhov/never-give-up}}} \\
    Mila, Universit\'e de Montr\'eal\\
    Allen Institute for AI
  \And
  \parbox[t]{0.36\textwidth}{\centering Hamish Ivison\thanks{Work performed while at the Allen Institute for AI}} \\
  University of Washington
  \AND
  \parbox[t]{0.36\textwidth}{\centering Nathan Lambert\footnotemark[2]} \\
  Trillium Labs
  \And
  \parbox[t]{0.36\textwidth}{\centering Aaron Courville} \\
    Mila, Universit\'e de Montr\'eal \\
    Canada CIFAR AI Chair
}

\begin{document}

\maketitle

\begin{abstract}
We demonstrate that training LLMs with RL does not improve performance equally across a dataset. RL shows large improvements on easy problems that an LLM is already good at solving, but small improvements on hard problems. 
We call this the \textit{Matthew Effect} in RL for LLMs, after the phenomenon of cumulative advantage from economics and network science summarized as ``the rich get richer''. 
The naive explanation is that hard problems require more compute to find a solution. We argue that modern RL methods are exacerbating the issue by wasting too much compute on easy problems and instead should dynamically reallocate how they use compute. 
We introduce \textit{Never Give Up} (NGU), a simple adaptive sampling method that keeps generating samples for a problem until one is correct. 
By leveraging asynchronous RL, this naturally uses fewer samples to filter out easy problems and allocates more compute to solving harder problems. 
We investigate the design choices that affect NGU, such as off-policy robustness, and develop a set of best practices. On the math benchmark Deepscaler, NGU improves performance per compute, especially on harder problems. On a recent coding task, Manufactoria, standard GRPO with a per-test reward fails to fully solve problems that have a range of easy and difficult tests. NGU iteratively improves, solving harder and harder tests, 
until it learns to fully solve coding problems.
\end{abstract}

\begin{figure}[h!]
    \centering
    \includegraphics[width=\linewidth]{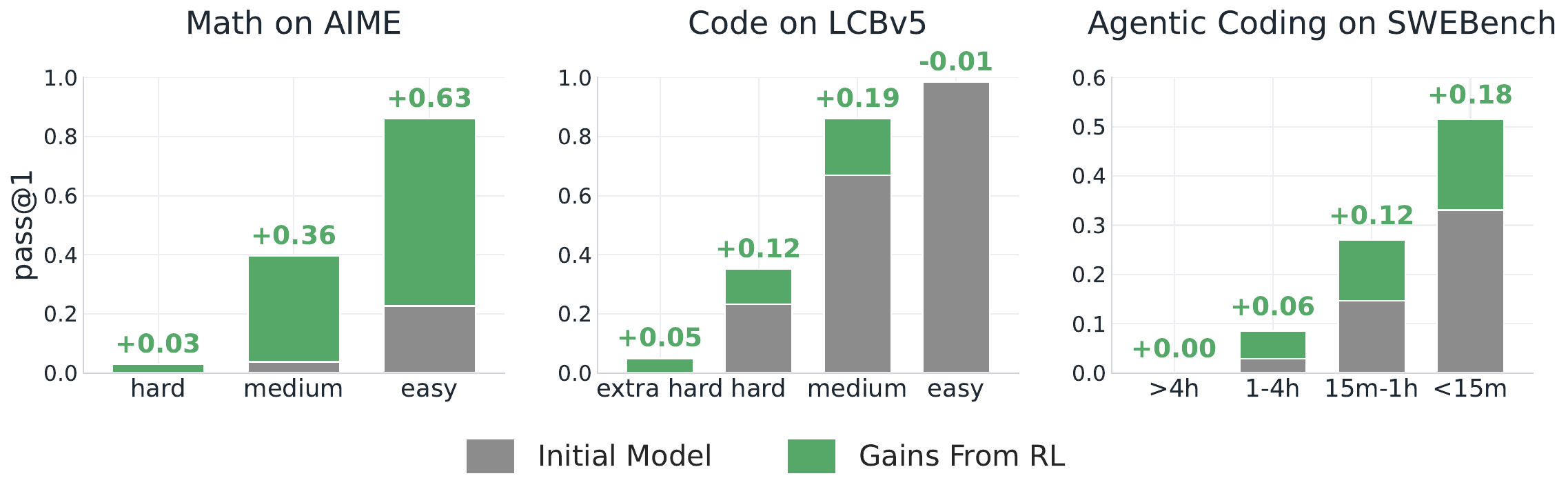}
    \vspace{-2em}
    \caption{\textbf{The Matthew Effect in RL for LLMs.} We evaluate how RL improves performance using three different open-source RL-trained models from three domains: Olmo 3.1 RL-Zero Math on AIME, DeepCoder on LCBv5, and DeepSWE on SWEBenchVerified. 
    We compare the RL-trained models to their initial models (Olmo 3 7B Base, Deepseek-Qwen-14B, Qwen3-32B), separating problems by difficulty.
    RL improves performance in proportion to the initial model's performance: easy problems improve the most and hard problems improve the least.
    }
    \label{fig:matthew-main}
\end{figure}

\section{Introduction}

Reinforcement learning (RL) is a standard method for post-training large language models~(LLMs) in the modern AI pipeline \citep{deepseek-ai_deepseek-v3_2025,olmo_olmo_2026}. 
Pre-training and mid-training imbue LLMs with strong priors and instruction-following abilities, but post-training with RL enables models to improve beyond their supervised data \citep{deepseek-ai_deepseek-r1_2025} in order to generalize to real world tasks \citep{chu_sft_2025}.

Practitioners generally assume that by training with RL on a range of easy to difficult problems, our models will learn to solve problems across the entire distribution. We find that this isn't true. In \autoref{fig:matthew-main}, we evaluate three different open-source RL-trained models from three different domains (math, coding, agentic coding) and find that easy problems receive a disproportionate amount of improvement compared to hard problems. Applying RL on LLMs shows a clear pattern of learning biased towards problems that the LLM was already good at. We connect this bias to a similar phenomenon in network science and economics, the Matthew Effect~\citep{merton_matthew_1968}, generally summarized as ``the rich get richer''. This work proposes \textbf{the Matthew Effect in RL for LLMs}. In \autoref{sec:matthew}, we define and demonstrate how RL training improves on problems in proportion to how easily the initial LLM can already solve them. We then make three major efforts towards elucidating this issue and enabling RL to solve harder problems.



\textbf{We find that inefficient compute allocation is one cause of the Matthew Effect} in \autoref{sec:causes}.
Intuitively, harder problems require more samples to reach a solution. But we argue that modern RL for LLM methods exacerbate the issue by assigning equal compute to all problems, regardless of difficulty. We show how naively sampling more completions for every prompt can add noise from spurious failures on easy problems and argue for dynamic compute allocation.

\textbf{We propose a dynamic sampling solution: Never Give Up (NGU) on unsolved prompts} in \autoref{sec:ngu}.  
NGU iteratively samples completions to a prompt and either (1) stops if a correct answer is found or (2) puts the prompt back in the queue to continue sampling. To avoid infinitely sampling an impossible prompt, we continue with some probability $p$ and otherwise give up. By leveraging modern asynchronous RL for LLMs \citep{noukhovitch_asynchronous_2024}, NGU uses fewer samples for easy problems and more samples for hard problems, naturally increasing their signal. We investigate important design choices for NGU and find that it helps to filter stale rollouts from the loss but including them in advantage calculations can improve performance.

\textbf{We empirically validate NGU on math and code RL at larger scales} in \autoref{sec:deepscaler} and \autoref{sec:manufactoria}. We apply NGU to a larger scale math task, and find that it outperforms strong RL baselines in solving the hardest problems. We then demonstrate an adapted Matthew effect for large prompt harnesses on a code generation task. Where standard RL stagnates, unable to pass the hardest coding tests, NGU dynamically allocates more compute to solve the coding problems and pass all tests. 

\section{The Matthew Effect in RL for LLMs}
\label{sec:matthew}

To demonstrate the Matthew Effect, we analyze three open-source models trained exclusively with RL in specific domains: Olmo 3.1 RL-Zero on math \citep{olmo_olmo_2026}, DeepCoder on code completion \citep{luo_deepcoder_2025}, and DeepSWE-Preview on agentic coding \citep{luo_deepswe_2025}. We group evaluation prompts by difficulty based on initial model performance: pass@1 on AIME (math), pass@1 on LCBv5 (code), and the human-time-based difficulty buckets provided in SWEBenchVerified.
As shown in \autoref{fig:matthew-main}, RL improvements scale with the model’s initial accuracy: tasks that are initially easier see larger gains, while harder tasks show more limited improvement. In other words, RL improves most where the model is already strong.
This pattern holds consistently across domains, models (Olmo 3, DeepSeek-Qwen2.5-R1-Distill, and Qwen 3), and datasets. This is a form of \textit{cumulative advantage}, and is analogous to the Matthew Effect in network science \citep{merton_matthew_1968}.
Originally observed in scientific citation patterns—where well-known researchers receive disproportionate credit for comparable work \citep{zuckerman_scientific_1977}—we adapt the concept to RL for LLMs:
\begin{quote}

\textbf{The Matthew Effect in RL for LLMs:} RL improves performance on a task in proportion to a model’s initial competence—making easy tasks easier while hard tasks often remain difficult.

\end{quote}

As this effect depends on the LLM's initial conditions, we can see it as a sort of primacy bias in RL \citep{nikishin_primacy_2022} where models are strongly impacted by their initial experiences. 
Previously, the RL primacy bias has focused on the issue of neural network plasticity loss when training from scratch \citep{nikishin_deep_2023}. In contrast, the Matthew effect requires pre-existing biases and therefore occurs in pretrained models with strong priors.



\section{Causes of the Matthew Effect}
\label{sec:causes}


The Matthew Effect may be partially explained by the mechanics of modern approaches to RL for LLMs, 
specifically GRPO \citep{shao_deepseekmath_2024} and its variants. These methods use an 
empirical group baseline \citep{kool_buy_2019,ahmadian_back_2024}: for each of the $N$ prompts, we sample $K$ completions and the advantage is computed as each completion's reward minus 
the group mean, $A_i = r_i - \frac{1}{K} \sum_i r_i$. The consequence is straightforward 
--- if all $K$ completions fail and receive zero reward, the prompt contributes no gradient. 
This \textit{signal loss} hypothesis \citep{xiong_reinforce-ada_2025} is an intuitive explanation; as 
noted by prior work \citep{qu_pope_2026}, RL can fail to improve on difficult problems simply 
because correct solutions are never sampled. The standard remedy is to increase $K$, raising 
the probability of sampling at least one correct solution for harder prompts 
\citep{hu_brorl_2025}. To investigate whether signal loss is indeed the primary driver of 
the Matthew Effect, we design an RL testbed for mathematical reasoning using a small-scale but high-quality dataset. 

\paragraph{Experimental Setup} We finetune Qwen 2.5 0.5B Instruct \citep{qwen_qwen25_2025} on math problems from GSM8k \citep{cobbe_training_2021}. 
To avoid issues with inaccurate data, we train on the cleaned and verified subset, GSM8k Platinum \citep{vendrow_large_2025}. 
We separate our evaluation into 4 distinct levels of problem difficulty: easy, medium, hard, and extra hard. 
To generate our evaluation set, we sample 1024 completions for each prompt using the initial model and extract 8 samples that correspond to four levels of difficulty: pass@1 of $25\%$ for easy, $10\%$ for medium, $5\%$ for hard, and $0\%$ for extra-hard.
Of extra-hard problems, 3/4 are completely unsolved in 1024 samples. 
We train our model with GRPO \citep{shao_deepseekmath_2024} sampling batches of $N$ prompts and $K$ completions-per-prompt. We leverage algorithmic refinements from recent works \citep{yu_dapo_2025,liu_understanding_2025} and train with off-policy asynchronous RL \citep{noukhovitch_asynchronous_2024} as it is the standard in large-scale post-training \citep{cursor_composer_2026,glm-5-team_glm-5_2026}. 
We filter out any prompt that receives no GRPO gradient (i.e., all-correct or all-incorrect completions) \citep{khatri_art_2025}. 
To maintain a constant batch size, we do active sampling \citep{olmo_olmo_2026} i.e. if a prompt is filtered, we sample more prompts until we have a full batch with non-zero GRPO gradient\footnote{This is the asynchronous analog to dynamic sampling used in synchronous RL \citep{yu_dapo_2025}}. 
We run all experiments for 3 seeds and report mean and standard deviation. See \autoref{app:gsm8k} for all experimental details.

\begin{figure}
    \centering
    \includegraphics[width=\linewidth]{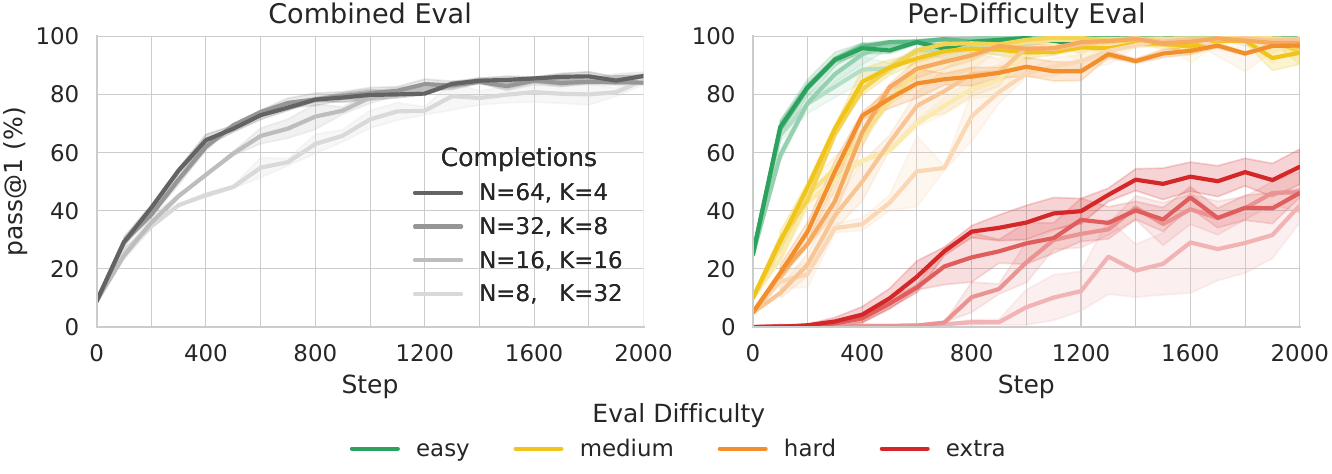}
    \vspace{-1em}
    \caption{\textbf{More Completions-per-prompt $K$ doesn't necessarily solve harder problems.} Under a fixed total batch size, using more completions per prompt (higher $K$, lighter lines) underperforms using 4 completions per prompt ($K=4$), even for harder problems.}
    \label{fig:gsm8k_baseline_passat1}
\end{figure}

\paragraph{Signal Loss?}
To test the signal loss hypothesis \citep{xiong_reinforce-ada_2025}, we examine the effect of increasing the number of completions-per-prompt $K$, intuitively reducing the chance of sampling all-incorrect responses. Specifically, we train with GRPO 
across four settings $K \in \{4, 8, 16, 32\}$, adjusting the number of prompts per batch
 $N \in \{64, 32, 16, 8\}$ to keep the total batch size $N \times K$ constant. As shown 
in \autoref{fig:gsm8k_baseline_passat1}, all settings exhibit the Matthew Effect: pass@1 
improves more rapidly on easier problems than on harder ones. Counterintuitively, the 
smallest setting $K=4$ performs best overall, which is clearly visible on the hardest 
problems (shown in red).
This result suggests that signal loss is not the primary driver of the Matthew Effect. 
Holding total batch size and training steps fixed, the expected number of times each 
problem is sampled remains constant regardless of $K$. What $K$ does control is the 
proportion of prompts filtered out due to a zero gradient --- that is, groups where all 
completions are either entirely correct or entirely incorrect. Increasing $K$ therefore 
does not increase the effective learning signal; it merely changes which prompts are 
discarded.

\begin{figure}
    \centering
    \includegraphics[width=\linewidth]{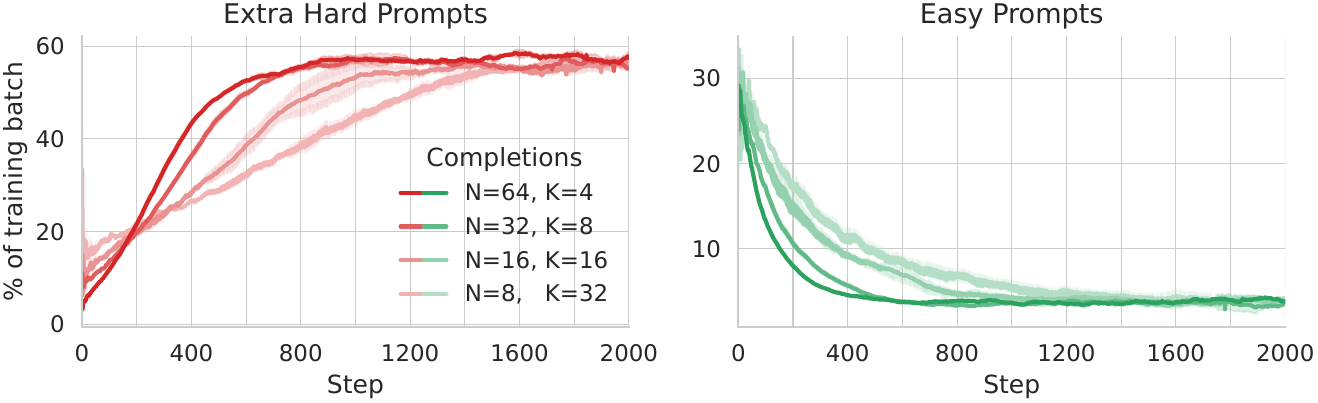}
    \vspace{-1em}
    \caption{\textbf{Fewer Completions-per-prompt $K$ can shift training batches towards harder prompts.} Using fewer completions per prompt (darker lines) results in more of the training batch comprising of hard training examples. This is due to more aggressive filtering of easy samples which enables focusing compute towards solving harder problems.}
    \label{fig:gsm8k_baseline_nonzero}
\end{figure}

Increasing the number of sampled completions, \(K\), has two competing effects. On one hand, larger \(K\) increases the probability of generating at least one correct completion for a difficult problem. On the other hand, it also increases the likelihood of sampling at least one \textit{incorrect} completion for an otherwise easy problem. As a result, smaller values such as \(K=4\) require relatively little compute to filter out easy prompts and continue searching for harder ones. In contrast, larger values such as \(K=32\) will end up including many more easy prompts in the training batch because it only requires a single incorrect completion among the 32 samples to be included in training. Consequently, larger \(K\) values allocate a greater fraction of training updates to problems that are already largely solved. To illustrate this effect, we partition the training set into four equal difficulty quartiles and track the composition of training batches over time. As shown in \autoref{fig:gsm8k_baseline_nonzero}, early in training, \(K=4\) contains fewer hard prompts because the model initially struggles to solve them, whereas \(K=32\) includes substantially more. However, by step 200, the smaller-\(K\) setting has filtered out easy prompts much more aggressively, resulting in training batches that are increasingly concentrated on harder examples.

\paragraph{Signal Efficiency} We propose an alternative interpretation: the \textit{signal efficiency} hypothesis for RL on LLMs.  The central issue is not merely whether enough completions are sampled for difficult problems but whether excessive compute is spent oversampling easy ones. From this perspective, RL training is fundamentally a problem of optimizing performance per unit of compute \citep{khatri_art_2025}. Standard GRPO always samples a fixed amount of completions \(K\), which creates an inherent inefficiency: when \(K\) is too large, substantial compute is wasted generating redundant completions for already-solved prompts; when \(K\) is too small, difficult problems are undersampled and therefore contribute little useful learning signal. This perspective also helps explain the effectiveness of recent multi-stage RL curricula that begin training with smaller \(K\) values and later transition to larger ones \citep{hu_brorl_2025}. Early low-\(K\) training can efficiently filter out easy prompts, while later high-\(K\) stages devote additional sampling budget to the increasingly difficult problems that remain unsolved. However, explicit curricula can be brittle and very sensitive to hyperparameters, we therefore aim to achieve signal efficiency using an adaptive, online method.

\section{Countering the Matthew Effect with Never Give Up}
\label{sec:ngu}

\paragraph{Not Resampling Easy Prompts}
The simplest solution has been to exclude problems that are too easy from being resampled later in training \citep{an_polaris_2025}. The issue is that this changes our training distribution to be harder and harder assuming that our model will never regress on previously-solved problems. In \autoref{app:extra_gsm8k} we find that our baseline training runs can perfectly solve a problem, but regress on it later in training when implementing this filtering strategy. So this approach can improve on the hardest problems at the expense of performance on easier problems.  

\paragraph{Never Give Up}
Our goal is therefore to keep sampling from our full training distribution but more quickly filter easy prompts while reallocating compute to hard prompts. We propose \textit{never give up}, a simple but effective way to leverage our asynchronous RL pipeline and reallocate compute to harder prompts. We first sample some small number of completions-per-prompt e.g. $K=4$. If all our completions are correct, we can quickly filter this problem as too easy. If all our completions are wrong, the standard approach is to give up and sample another problem. Instead, we \textit{never give up} (NGU) on solving this problem and sample $K$ more completions, continuing until we solve the prompt. As certain problems may be too difficult, we actually continue sampling with probability $p_{\text{NGU}}$ and give up on the problem with probability $1-p_{\text{NGU}}$. This creates a geometric distribution of the total number of samples, allowing us to occasionally sample many completions on difficult prompts but keeping the average number of samples at $\frac{K}{1-p}$. See pseudocode for NGU in \autoref{app:pseudocode}.

\begin{figure}
    \centering
    \includegraphics[width=\linewidth]{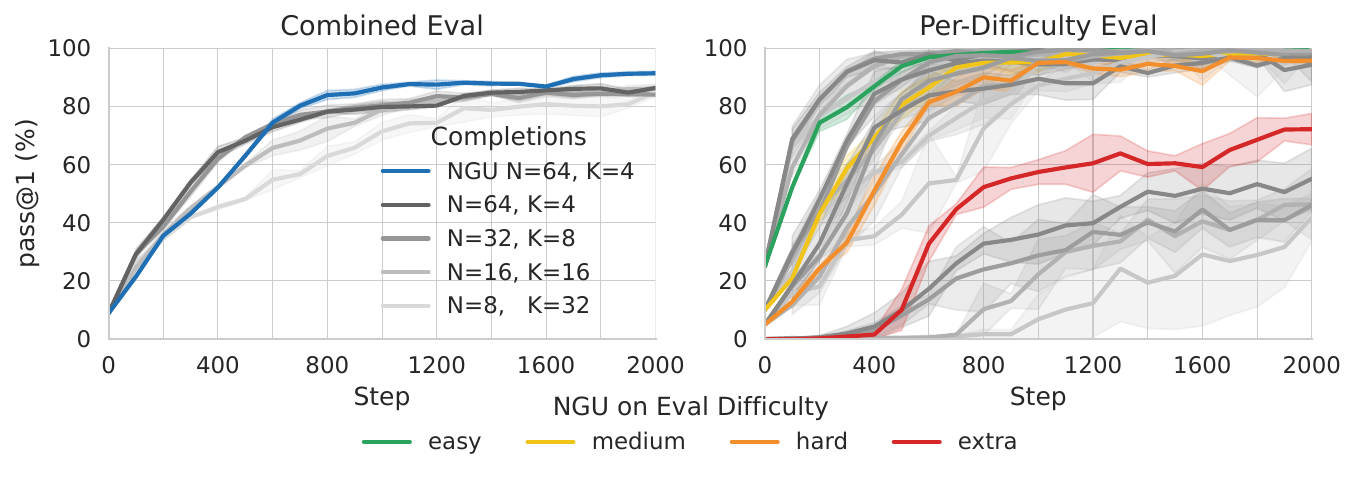}
    \vspace{-2em}
    \caption{\textbf{Never Give Up improves on the hardest problems.} Using NGU improves overall pass@1 (blue line on left), especially improving performance on the hardest subset of samples (red line on right). Baseline GRPO results from \autoref{fig:gsm8k_baseline_passat1} are shaded in grey.}
    \label{fig:gsm8k_ngu_passat1}
\end{figure}

\begin{figure}
    \centering
    \includegraphics[width=\linewidth]{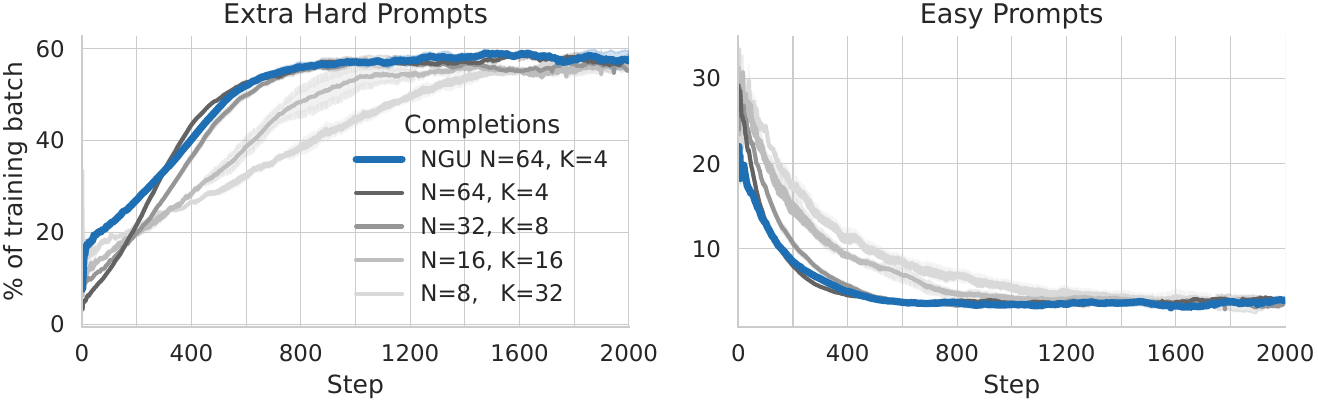}
    \vspace{-2em}
    \caption{\textbf{Never Give Up achieves near pareto-optimality for training on hard prompts.} Using NGU results in the highest proportion of difficult samples and lowest proportion of easy samples across all training steps, suggesting it gets the benefits of both small and large $K$ through its dynamic allocation. Baseline GRPO results shaded in grey (see \autoref{fig:gsm8k_baseline_nonzero}). }
    \label{fig:gsm8k_ngu_nonzero}
\end{figure}

We re-run our previous $N=64, K=4$ baseline but continue sampling with $p_{\text{NGU}} = 0.95$. As shown in \autoref{fig:gsm8k_ngu_passat1}, NGU outperforms all previous baselines with its pass@1 gains coming specifically on the extra hard problem subset. Looking at what prompts NGU trains on in \autoref{fig:gsm8k_ngu_nonzero}, we see that NGU gets the best of both large $K$ and small $K$. Early in training, NGU has as many hard prompts in the batch as $K=32$ but later in training it has as many as $K=4$, achieving close to the pareto-optimal amount of hard prompts in the training batch. Similarly, NGU achieves the best of both $K=4$ and $K=32$ for filtering easy prompts. Key to NGU's success is an asynchronous RL infrastructure, so that quickly filtered easy prompts are replenished by harder prompts, therefore reallocating compute from easy to harder problems. We demonstrate the necessity of async RL for NGU and provide a detailed comparison to previous sync RL work \citep{xiong_reinforce-ada_2025} in \autoref{app:reinforce-ada}


\begin{figure}
    \centering
    \includegraphics[width=\linewidth]{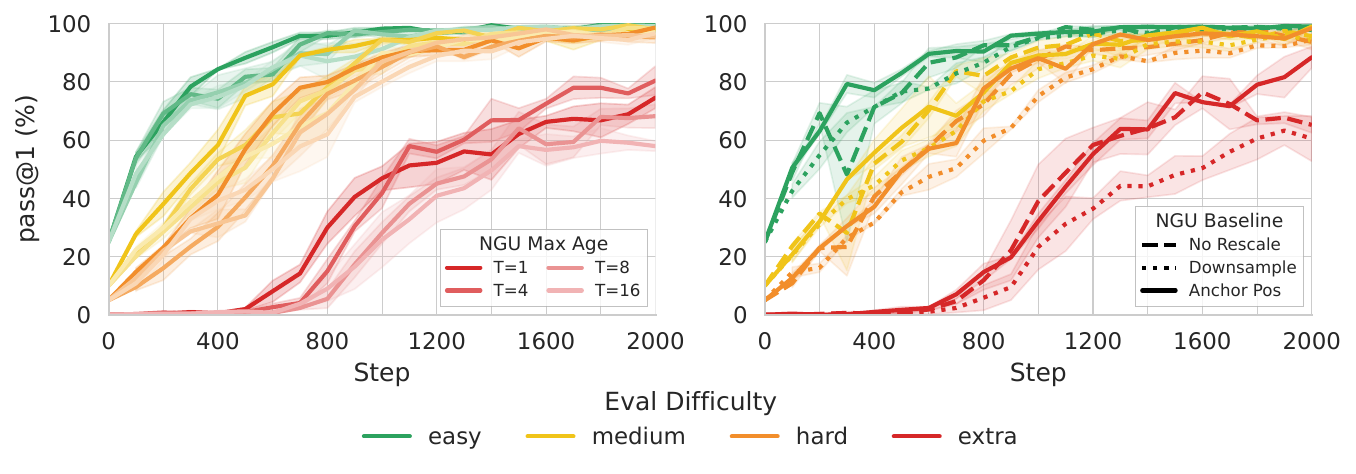}
    \vspace{-2em}
    \caption{\textbf{Never Give Up requires filtering of stale/off-policy completions, but can use them for an improved GRPO baseline.} We measure performance across training and across samples of different difficulties. Using older completions ($T>4$ steps off-policy) results in deteriorating performance (left), but these completions can be used for baselining. Using them for anchoring the positives outperforms other baselining approaches, especially downsampling (right).}
    \label{fig:gsm8k_ngu_age_passat1_baseline_passat1}
\end{figure}

\paragraph{NGU: Maintaining Previous Completions}
Another advantage of NGU over plain GRPO with $K=4$ is that we can get a richer feedback signal by collecting a larger group size for harder problems. Standard sampling with $K=4$ may occasionally sample the right answer for a very difficult prompt but the advantage can be at most $1 - \frac{1}{4}$ in our GRPO setup, even if the many previous samples of this prompt were unsuccessful. NGU resolves this problem by maintaining previous completions and, once a correct completion is found, forming one big GRPO group with all previous completions. A large group size with few correct answers greatly increases the advantage of the right answers, which allows for updating rare correct answers with a larger gradient. The disadvantage of this approach is that asynchronous NGU resampling can keep stale, off-policy negative completions from previous steps, which some recent work suggests can be harmful \citep{roux_tapered_2025,fu_areal_2025}. To counter this, we keep track of the age of previous NGU completions and maintain only those that are less than $T$ steps old. We run our experiment across $T = \{1, 4, 8, 16\}$ where $T=1$ only maintains the current $K=4$ completions. As shown in \autoref{fig:gsm8k_ngu_age_passat1_baseline_passat1}, the performance increases from $T=1\to 4$ but then decreases with $T=8,16$. 

\paragraph{NGU: Maintaining Counts}
Since it is optimal to only maintain previous completions with staleness of less than $T=4$, it would be useful to still somehow leverage older completions that we filter from our training. An intuitive idea is to leverage all previous completions' reward in calculating our true GRPO baseline $\bar r_{\text{NGU}} = \frac{1}{K*NGU} \sum_i r_i$, even if we don't use all those completions in our update. The issue is that this will de-center the average advantage for a GRPO group as we will use $\bar r_{\text{NGU}}$ for a baseline but filter out some negative completions and their advantages for being too old.\footnote{Note that positives are new and only negatives can be stale, as we stop sampling once we achieve a positive} We propose an intuitive novel solution: maintain the advantage of the positive completions and evenly rescale the advantages of the negative completions to $\frac{n_+}{n_-}(1-\bar r)$ where $\frac{n_+}{n_-}$ is the ratio of positives to remaining negatives. This guarantees that, as before, the sum of advantages for our update group is 0. We call this \textit{anchoring the positives} and compare it to two baselines. The first is simply to maintain a de-centered sum of advantages i.e. no rescaling. The second, proposed by \citet{xiong_reinforce-ada_2025}, is to downsample our negative completions so that there are an equal number of positive and negative completions then rescale advantages by $\frac{1}{\bar r_\text{NGU}}$ \footnote{This approach is reminiscent of MaxRL \citep{tajwar_maximum_2026} which calculates GRPO advantage as $(r_i - \bar r) / (\bar r)$}. We compare methods in \autoref{fig:gsm8k_ngu_age_passat1_baseline_passat1} and find that anchoring the positives is generally better, improving on the hardest prompts. In particular, downsampling and getting rid of useful negatives appears quite harmful to performance. In \autoref{app:extra_gsm8k}, we also demonstrate that our NGU baseline with stale completions' rewards is better than a standard GRPO baseline using only completions we train on.

\section{Math: Scaling Up NGU}
\label{sec:deepscaler}

We now aim to validate our method and design choices at a larger scale, on a harder task. 

\paragraph{Experimental Setup} We train Qwen 3 4B-base \citep{yang_qwen3_2025} on a 10k subset of Deepscaler math \citep{luo_deepscaler_2025} following previous work \citep{li_jointly_2025}, and evaluate on difficult problems from recent math contests, AIME 2025 and BRUMO Nov 2025 \citep{dekoninck_beyond_2026}. We divide our combined evaluation prompts into difficulty buckets based on our initial model's pass@64: hard includes all samples with pass@64=0, medium averages $2.9\%$, easy averages $46.1\%$. 

We run our baselines varying completions-per-prompt $K \in \{16, 32, 64\}$ and maintain the same batch size with corresponding $N \in \{8, 4, 2 \}$. We compare to $K=16, N=8$ with NGU varying $p_\text{NGU} \in \{0.5, 0.75, 0.875\}$ to give theoretical average sampling ranges of $K = 32,64,128$ respectively. For a fair comparison, we compute-match NGU by running all experiments for approximately 120 H100 hours regardless of the number of steps and run three seeds per setting. 

\begin{figure}
    \centering
    \includegraphics[width=0.8\linewidth]{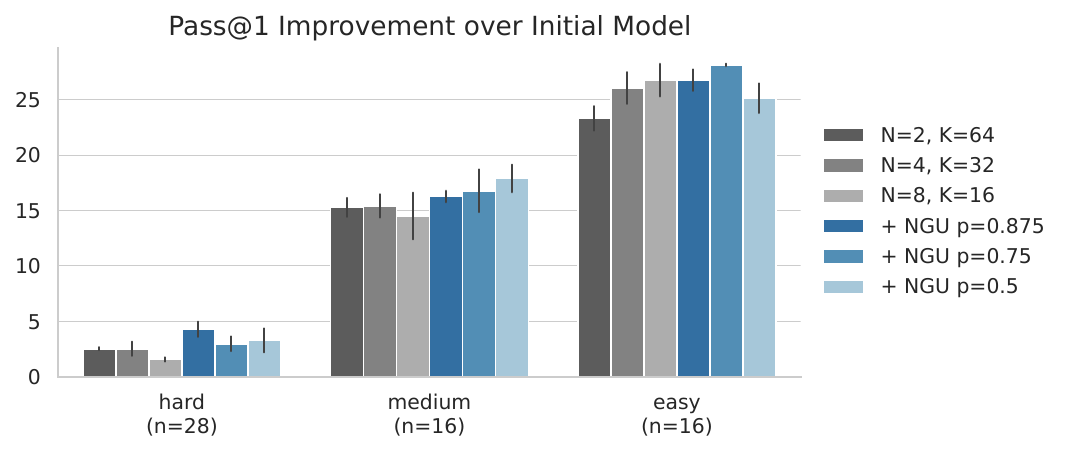}
    \caption{\textbf{NGU can improve the performance trade-off between hard/easy samples.} We train on Deepscaler with different values of $K$ and eval on AIME and BRUMO 2025. NGU outperforms all options, especially on the hardest problems. Varying NGU $p$ demonstrates a trade-off between improving on hard and easy evals, much like varying $K$. We show mean $\pm$ std dev across three seeds.}
    \label{fig:deepscaler_ngu}
\end{figure}

\paragraph{NGU improves the easy/hard performance trade-off} In \autoref{fig:deepscaler_ngu} we find that increasing $K$ in our baseline trades off performance between easy and hard problems, similar to results on GSM8k. Varying $p_\text{NGU}$ better trades off performance, improving more on hard problems without sacrificing as much performance on easy problems. As in GSM8k, we attribute NGU's success to implicitly having more difficult prompts in the training batch, which we show in \autoref{fig:deepscaler_prompt_ratios} in \autoref{app:extra_deepscaler}. 

\paragraph{NGU outperforms curriculum learning}
We compare against a simple curriculum learning baseline that sets $K$ per prompt using a model's initial performance on that prompt. Fixed, larger $K$ for hard prompts successfully reallocates compute and matches NGU's performance on the hardest subset. But it degrades performance on easy problems, demonstrating that difficulty estimation is dynamic and can change over training. In \autoref{app:extra_deepscaler}, we show results across eval datasets in \autoref{tab:deepscaler_by_dataset} and results across difficulty levels in \autoref{tab:deepscaler_improvement_pass_at_1}. 

\section{Coding: From Prompt Difficulty to Test Difficulty}
\label{sec:manufactoria}

Code generation, unlike math, can contain levels of difficulty not just between prompts but between tests within a single prompt. We demonstrate the benefit of NGU on a difficult coding task where standard RL doesn't learn to pass the hardest coding tests.

\paragraph{Experimental Setup} We experiment on a recent coding benchmark, Manufactoria \citep{manufactoria}, a classic Flash game in which players build automated factories to sort robots based on their colored tape patterns. The underlying logic resembles constructing finite-state automata or tag systems and the benchmark allows for complex OOD training that evaluates RL's ability to generalize. Following the original benchmark \citep{sun_rl_2025}, we train Qwen3 4B Instruct \citep{yang_qwen3_2025} on a set of diverse and challenging coding problems, then evaluate on a test set that measures generalization. Each problem contains between 14 and 30 test cases, covering a range of difficulties. The final test cases in each example are frequently quite hard, making it feasible to solve a few test cases, but difficult to pass all tests. See all experimental details in \autoref{app:manufactoria}.

\begin{figure}
    \centering
    \includegraphics[width=\linewidth]{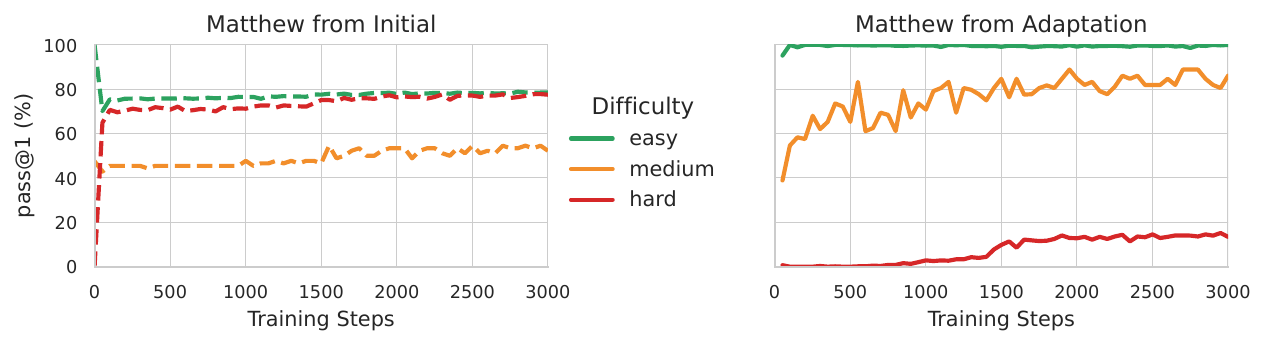}
    \vspace{-1em}
    \caption{\textbf{The Harness-Aware Matthew Effect}. Performance on easy, medium, and hard problems over training. We can determine problem difficulty based on the model's own pass rate either at the start of training (left) or after 100 steps when the model has adapted to the novel prompt/harness (right). Using the adapted difficulty ratings leads to a clear demonstration of the Matthew effect.}
    \label{fig:manufactoria_matthew}
\end{figure}

\paragraph{The Harness-Aware Matthew Effect}
To evaluate the Matthew effect on Manufactoria, we take 128 samples from our initial model, Qwen 3 4B Instruct, to assign a difficulty level for each prompt. We then train our model with GRPO following \citet{sun_rl_2025}, and plot the performance over each difficulty level over time. As shown in \autoref{fig:manufactoria_matthew} (left), we see a surprising lack of a Matthew effect. This can be reasonably explained by Manufactoria being both a novel task and using a long and novel prompt in order to make the output conform to requirements. The model is not well adapted to its prompt at initialization. Early RL steps are therefore more about adapting the model's capabilities to the prompt, than a reflection of true improvement on the task \citep{liu_understanding_2025}. Training shows two clear phases of (1) adapting to the task and prompt then (2) learning to solve the task with the specified prompt.
For that reason, we propose to measure our Matthew effect using the model \textit{after} prompt/harness adaptation has completed, around step 100. Re-classifying our problem difficulty after adaptation, in \autoref{fig:manufactoria_matthew} (right), we again see a clear Matthew effect appear in our results. We therefore note the modified harness-aware Matthew effect for RL in LLMs: 

\begin{quote}
\textbf{The Harness-Aware Matthew Effect in RL for LLMs:} RL improves an LLM's performance on a task in proportion to the model's \textit{early} performance on the task \textit{after accounting for adaptation to a prompt or harness}.
\end{quote}

\begin{figure}
    \centering
    \includegraphics[width=\linewidth]{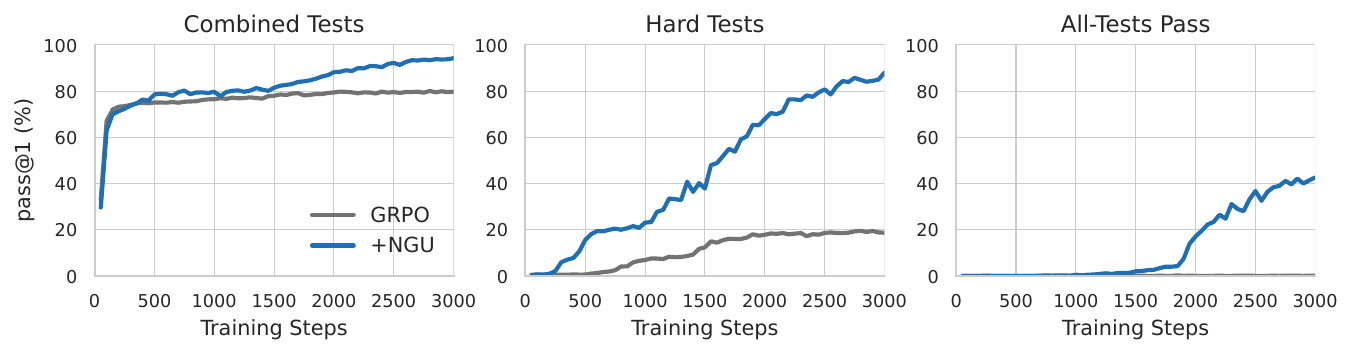}
    \vspace{-1em}
    \caption{\textbf{Never Give Up enables solving the hardest tests on coding problems.} NGU enables continuously improving on all tests even as GRPO plateaus (left). The difference between the two methods is on hard tests (center). NGU improves on the hardest tests and learns to pass all of a problem's tests, while GRPO remains unable to pass all tests on even a single example (right).}
    \label{fig:manufactoria_ngu}
\end{figure}


\paragraph{Signal Efficiency in Tests}\citet{sun_rl_2025} observed that a baseline GRPO run using per-test rewards was unable to fully solve the task. Although the model learned to solve roughly \(80\%\) of all coding tests, it rarely learned to pass all tests for any given problem. We interpret this phenomenon through the lens of the \textit{signal efficiency} hypothesis. As shown in \autoref{fig:manufactoria_matthew} (right), a standard RL run almost always solves the easiest tests, rarely exceeds \(20\%\) success on the hardest tests, and derives most of its reward variation from inconsistently solving medium-difficulty tests. Consequently, the majority of the training signal is allocated to repeatedly revisiting partially solved tests rather than driving progress on the hardest unresolved ones. In this regime, compute is spent inefficiently on fluctuations around already-near-solved behaviors instead of generating an informative learning signal for full task completion. In contrast, NGU adaptively concentrates sampling on trajectories that improve over previous solutions. By continuing to sample until harder tests are solved, it allocates compute toward the hardest, most informative test cases 
rather than repeatedly oversampling easy or unstable intermediate cases.


We now run NGU on Manufactoria with the same settings as the baseline, adding only $p_\text{NGU}=0.95$, shown in \autoref{fig:manufactoria_ngu}. As before, we confirm the main result of \citet{sun_rl_2025} that standard GRPO cannot learn to solve all tests and find the reason is a stagnation in solving the hardest subset of tests (according to our harness-aware difficulty levels). In contrast, though NGU is initially slightly worse on the overall combined test pass-rate, it continues improving on hard tests and eventually transfers this capability into fully solving all tests and passing full problems.

\begin{wrapfigure}{r}{0.5\textwidth}
    \centering
    \includegraphics[width=\linewidth]{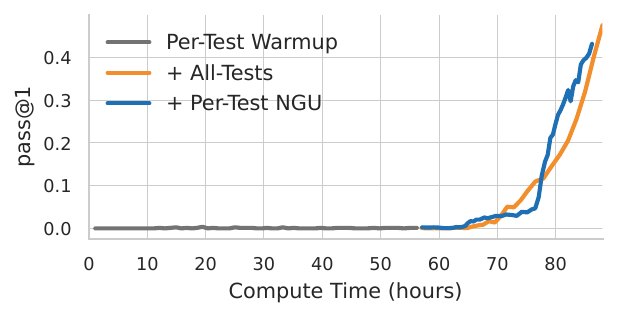}
    \caption{\textbf{Never Give Up can recover from a suboptimal model overfit to per-test reward}. 
    NGU with per-test rewards performs similarly to standard RL with all-test pass reward.
    }
    \label{fig:manufactoria_resume_compute_time}
\end{wrapfigure}

\paragraph{Comparison to the Primacy Bias in RL}
The only method that \citet{sun_rl_2025} found that could solve Manufactoria was to first train with a reward \textit{per-test} and, once performance has plateaued, restart training with a reward only if \textit{all-tests} pass. The \textit{all-tests} reward can avoid issues of signal efficiency but it isn't possible to train with it from the start as the initial model never manages to pass all tests. To directly compare all-tests reward from \citet{sun_rl_2025} with per-test reward using NGU we start both runs from the final GRPO baseline checkpoint after 3000 steps. In \autoref{fig:manufactoria_resume_compute_time} we compare both methods across compute time as all-tests reward runs slower per step due to larger amounts of filtering. We see that per-test reward with NGU approximately matches all-tests reward in compute efficiency. Interestingly, this result also demonstrates that RL for LLMs can train for 3000 steps with a suboptimal objective and then later recover when given a good training signal. In contrast, the primacy bias in RL \citep{nikishin_primacy_2022} found that RL models trained from scratch will fail to recover from initial bad trajectories, due to plasticity loss. 
This result implies that RL for LLMs does not suffer the same plasticity loss and, despite being a form of primacy bias, the Matthew effect is distinct from plasticity.

\section{Related Work}

Many works in RL for LLMs have tackled solving harder problems and recent methods have usually focused on leveraging external information \citep{qu_pope_2026, wu_learn_2026}, modifying the GRPO loss \citep{tajwar_maximum_2026}, or improving test-time search \citep{hubert_olympiad-level_2026}. Another line of work has focused on creating an explicit curriculum of difficulty, either by removing easy problems \citep{an_polaris_2025}, curating an explicit dataset \citep{luo_deepscaler_2025}, or adjusting the sampled data distribution~\citep{shi_efficient_2026, qu_can_2026,zheng_act_2025}. As NGU does not explicitly shift the data distribution but implicitly adapts it to improve performance-for-compute, these approaches are orthogonal and complementary to ours. The most similar method to NGU is Reinforce-Ada-Seq-Positive \citep{xiong_reinforce-ada_2025}, aimed at solving the issue of undersampling for hard prompts i.e. \textit{signal loss}, as opposed to reallocating compute i.e. \textit{signal efficiency}. This difference is driven by their use of synchronous RL, which removes any compute efficiency benefits of NGU. We delve into the differences and note the necessity of asynchronous RL to our method in \autoref{app:reinforce-ada}.
RL work prior to LLMs tackled highly difficult scenarios through the lens of exploration in sparse reward tasks \citep{bellemare_unifying_2016}.
Methods generally focused on identifying promising states \citep{ecoffet_go-explore_2021} and using auxiliary losses to guide exploration \citep{badia_never_2020}. As our current RL for LLM setups are simple contextual bandits, not MDPs, we find NGU currently sufficient to improve performance in this setting. We provide an extended related work in \autoref{app:related}.

\section{Conclusion}

In this work, we have highlighted \textit{the Matthew Effect} in RL for LLMs, and shown that the standard approach results in disproportionately poor performance on the hardest problems. 
This also demonstrates how simple scalar values may not be sufficient for accurate evaluations of LLMs, and we should move towards dense evaluation signals. 
Our proposed solution, Never Give Up, represents a simple method for better allocating compute that allows substantial improvement on difficult tasks. Future work could examine more complex multi-step, agentic environments. As the feasibility of RL always comes down to a tradeoff of performance and compute~\citep{khatri_art_2025}, future scaling of RL must continue to find ways to efficiently solve and progress on increasingly hard tasks.

\subsection*{Acknowledgements}

MN thanks Shengyi Costa Huang for all the feedback, advice, and direction. MN thanks Samuel Lavoie, Dzmitry Bahdanau, and Finbarr Timbers for many helpful discussions and Yiyou Sun for help with implementing Manufactoria and reproducing results. MN was supported by the Fonds de recherche du Qu\'ebec - Nature et Technologies. This research was supported by CIFAR, IVADO, and International Joint Research Project — KAIST NAIRL. We thank the Beaker and infrastructure teams at Ai2 for their support with Ai2 cluster. This material is based upon work supported by the National Science Foundation under Award No. 2413244 and research was enabled in part by support and compute provided by Mila (mila.quebec), Calcul Quebec (calculquebec.ca), and the Digital Research Alliance of Canada (alliancecan.ca).

\bibliography{references,references-extra}

\newpage
\appendix

\section{Extended Related Work}
\label{app:related}

The problem of scaling RL performance for compute has been investigated across multiple different dimensions. \citet{khatri_art_2025} perform an extensive search across different algorithmic decisions. \citet{he_justrl_2025} conduct a smaller scale study and focus on simple rules that improve performance for small models trained on Math. Recently \citet{liu_part_2025} dive into the interactions between dataset composition and RL algorithms, noting best practices at their intersection. These papers all perform important investigations and we leverage several of their recommendations in our algorithmic choices. 

Our specific goal of solving harder problems has long been a topic of interest in RL and our proposed method has similarities to many prior works. In bandits, \citet{audibert_best_2021} aim for best-arm selection and propose successive rejection as the basis for improved exploration. In deep RL, \citet{jiang_prioritized_2021} aim to solve harder tasks and use TD-error to guide a curriculum towards harder levels of training. In recent RL for LLMs,
GVM-RAFT \citep{yao_optimizing_2025} also proposes to vary group sizes as a form of curriculum, used in an EM method that leverages SFT to approximate RL. Their method relies on predicting the correct group size using gradient norms and also is not particularly effective when applied to GRPO. Another recent method, GRESO \citep{zheng_act_2025}, also seeks to improve performance for compute but focuses on avoiding signal loss, as in \citet{xiong_reinforce-ada_2025}. They use an online curriculum, similar to No-Positive Resampling \citep{an_polaris_2025} but with a probabilistic filtering of problems, separately tracking too easy and too difficult problems. Compared to NGU, GRESO requires many more hyperparameters to define their curriculum and may not improve performance on the hardest examples as it explicitly samples fewer difficult problems to avoid all-zero-reward groups. Additionally, these previous works leveraging online curricula do not investigate code RL where a single problem can contain tests of variable difficulties.

\section{Limitations}

The intuition behind Never Give Up is that our RL training can reallocate compute by quickly filtering easy problems. This implicitly assumes a range of difficulties for our RL training distribution with a sufficient amount of easy tasks that need filtering. As shown in \autoref{fig:deepscaler_difficulty} in \autoref{app:deepscaler}, our Deepscaler task does indeed have a range of difficulties that leans towards difficult prompts, but has a reasonable amount of easy and medium prompts as well. If a task leans more heavily towards difficult problems, Never Give Up will likely not be effective. This is because NGU's iterative sampling process may take longer to finish a whole group of completions compared to explicitly sampling larger $K$ from the start. This means finished groups of NGU completions are generally more asynchronously off-policy and therefore provide a worse training signal \citep{openai_dota_2019}, especially for negative samples \citep{roux_tapered_2025}. For this reason, we also do not recommend setting $K=1$ and fully controlling batch size with $p_{NGU}$ as it leads to more off-policy training.

\section{Extra Results}

\subsection{Comparison to Reinforce-Ada}
\label{app:reinforce-ada}

Reinforce-Ada-Seq-Pos is the most similar method to Never Give Up in the current literature, but this work notably differs from Reinforce-Ada \citep{xiong_reinforce-ada_2025} in both method and motivation. 

\paragraph{Signal Loss vs Signal Efficiency} Reinforce-Ada methods aim to solve a different issue, \textit{signal loss}: when all completions to a prompt have the same reward, giving zero GRPO gradient. Signal loss also takes issue with a prompt giving all \textit{positive} completions, and therefore no gradient. So other methods, such as Reinforce-Ada-Seq-Balance also aim to increase sampling on very easy problems in order to get an unlikely negative sample. In contrast, our proposed \textit{signal efficiency} sees unlikely positive samples as a useful signal but unlikely negative samples as noise. 

\paragraph{Reinforce-Ada-Seq-Pos vs Never Give Up}
Though both methods continue sampling until a positive completion is achieved, Reinforce-Ada-Seq-Pos differs empirically from NGU in three important ways:

\begin{enumerate}
    \item No "giving up". 
    \citet{xiong_reinforce-ada_2025} do not stop sampling until a positive is reached, equivalent to NGU with probability $p=1$. Therefore, this method stalls on any prompt that is unsolvable by the model. 
    \item Synchronous RL. \citet{xiong_reinforce-ada_2025} propose continuing to sample in a synchronous RL framework. It therefore doesn’t benefit from filtering positives quickly as the batch still waits for all (hard) completions to finish before sampling another prompt.
    \item Downsampling completions. \citet{xiong_reinforce-ada_2025} generally strives for a balanced batch containing an equal number of positives and negatives. Therefore Reinforce-Ada-Seq-Pos discards negative completions so that they match the number of positive completions.
\end{enumerate}

We ablate these three design decisions on the GSM8k task and show results in \autoref{tab:ngu_v_reinforce-ada-seq-pos}. We find that $p=1.0$ leads to stalling and is empirically quite bad. NGU $p=0.75$ with sync RL underperforms baseline GRPO, demonstrating the necessity of asynchronous infra for NGU to be effective. And finally, we find that downsampling is less effective than our choice of anchoring positives and, as shown in \autoref{fig:gsm8k_ngu_anchorpos_v_none_passat1}, it is also worse than doing nothing.

\begin{table}[]
    \centering
    \caption{\textbf{Ablating the differences between NGU vs Reinforce-Ada-Seq-Pos on GSM8k.} Each design decision behind Never Give Up is shown to be crucial to performance: giving up with probability $p$, asynchronous RL, anchoring positives for rescaling reward baseline.}
    \label{tab:ngu_v_reinforce-ada-seq-pos}
  \begin{tabular}{lllll}
  \toprule
   & easy & medium & hard & extra \\
  \midrule
Baseline GRPO & \textbf{100.0 ± 0.0} & \textbf{99.9 ± 0.1} & \textbf{99.6 ± 0.2} & 47.1 ± 6.4 \\
+ NGU $p=1.0$ & 83.9 ± 0.7 & 61.2 ± 2.3 & 36.3 ± 3.3 & 2.7 ± 1.0 \\
  + NGU $p=0.75$ + sync RL & 99.0 ± 0.5 & 97.3 ± 0.5 & 93.8 ± 2.3 & 14.8 ± 5.2 \\
  + NGU $p=0.75$ + async RL + downsampling & 99.5 ± 0.3 & 96.7 ± 0.7 & 95.4 ± 0.9 & 63.9 ± 5.5 \\
  + NGU $p=0.75$ + async RL + Anchor Pos & \textbf{100.0 ± 0.0} & 99.2 ± 0.4 & 98.6 ± 0.3 & \textbf{87.9 ± 2.2} \\
  \bottomrule
  \end{tabular}
\end{table}

\subsection{GSM8k}
\label{app:extra_gsm8k}

\begin{figure}[h!]
    \centering
    \includegraphics[width=\linewidth]{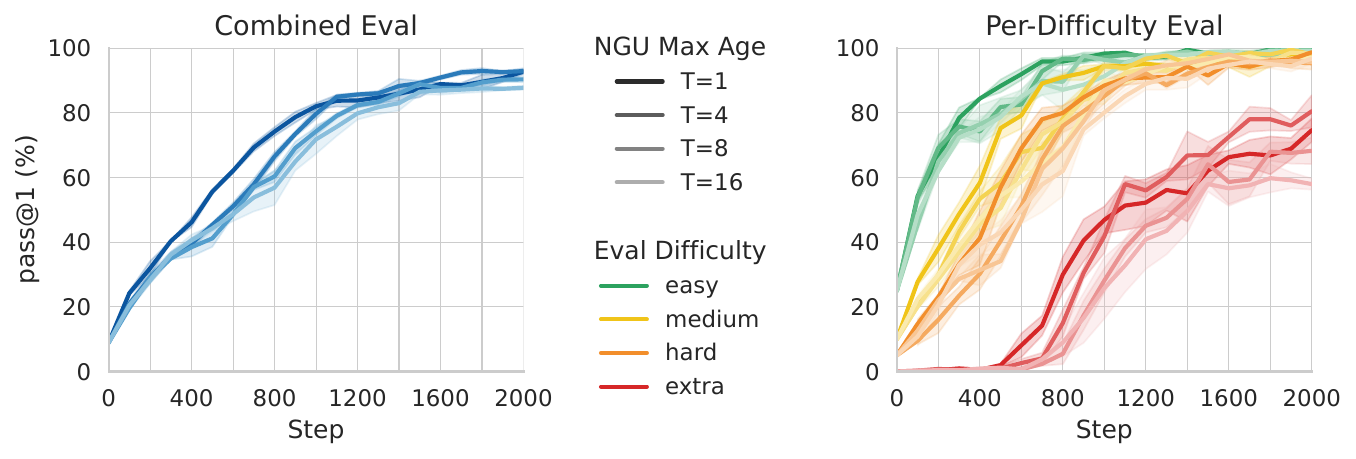}
    \vspace{-1em}
    \caption{\textbf{Never Give Up must filter previous completions if they are too stale/off-policy.} Both overall (left) and per-difficulty (right), using more stale completions (lighter lines) lowers performance when using NGU.}
    \label{fig:gsm8k_ngu_age_passat1}
\end{figure}

In \autoref{fig:gsm8k_ngu_age_passat1}, we show full overall and per-difficulty results when varying the maximum age of completions used in NGU.

\begin{figure}[h!]
    \centering
    \includegraphics[width=\linewidth]{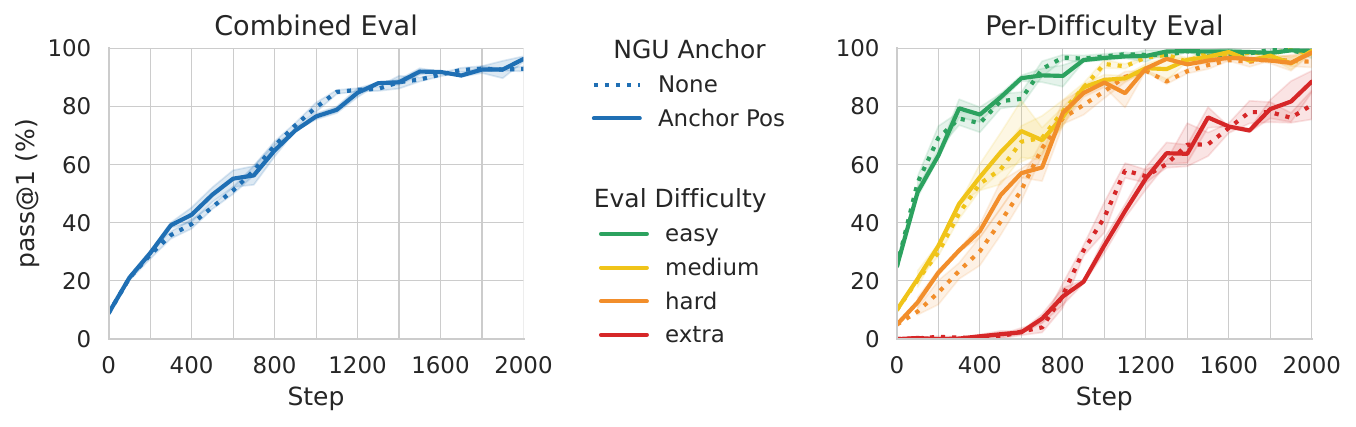}
    \vspace{-1em}
    \caption{\textbf{Never Give Up can leverage filtered completions for an improved GRPO baseline.} We compare using stale completions with Anchor Pos rescaling to not using filtered completions for the GRPO baseline and find previous, filtered completions can be beneficial.}
    \label{fig:gsm8k_ngu_anchorpos_v_none_passat1}
\end{figure}

In \autoref{fig:gsm8k_ngu_anchorpos_v_none_passat1}, we show full overall and per-difficulty results for different ways of rescaling advantages in the presence of filtered negatives. We compare not rescaling, \citet{xiong_reinforce-ada_2025} downsampling, and our ``anchoring the positives''.

\begin{figure}[h!]
    \centering
    \includegraphics[width=\linewidth]{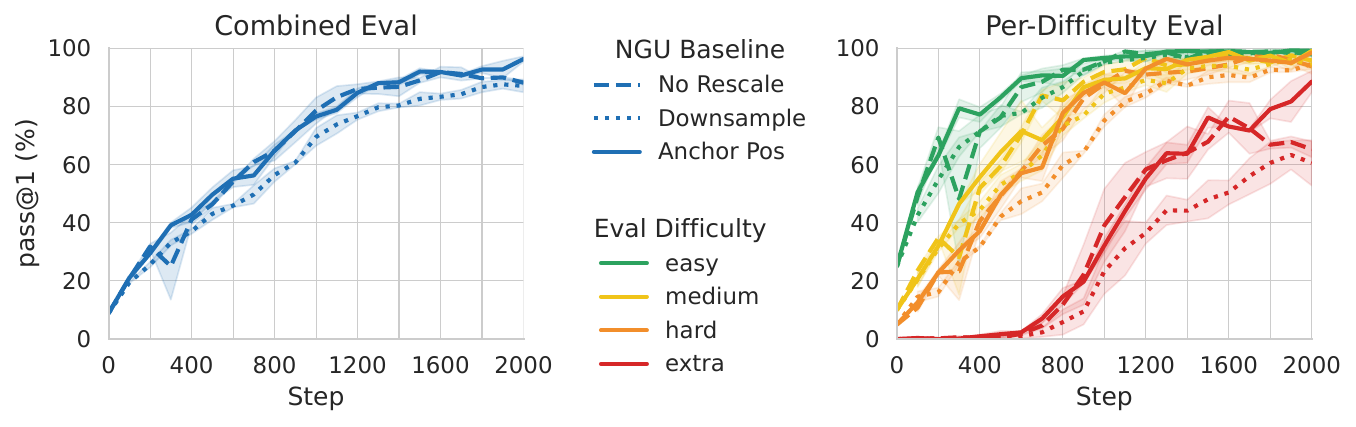}
    \vspace{-1em}
    \caption{\textbf{Never Give Up can best leverage filtered completions for an improved GRPO baseline by rescaling rewards.} We ablate three methods for using stale completions. Downsampling underperforms both other approaches, and anchoring the positive completions performs best at the end of training.}
    \label{fig:gsm8k_ngu_filtered_rescaling_passat1}
\end{figure}

In \autoref{fig:gsm8k_ngu_filtered_rescaling_passat1}, we show full overall and per-difficulty results when varying the baselining method used in NGU.

\begin{figure}[h!]
    \centering
    \includegraphics[width=\linewidth]{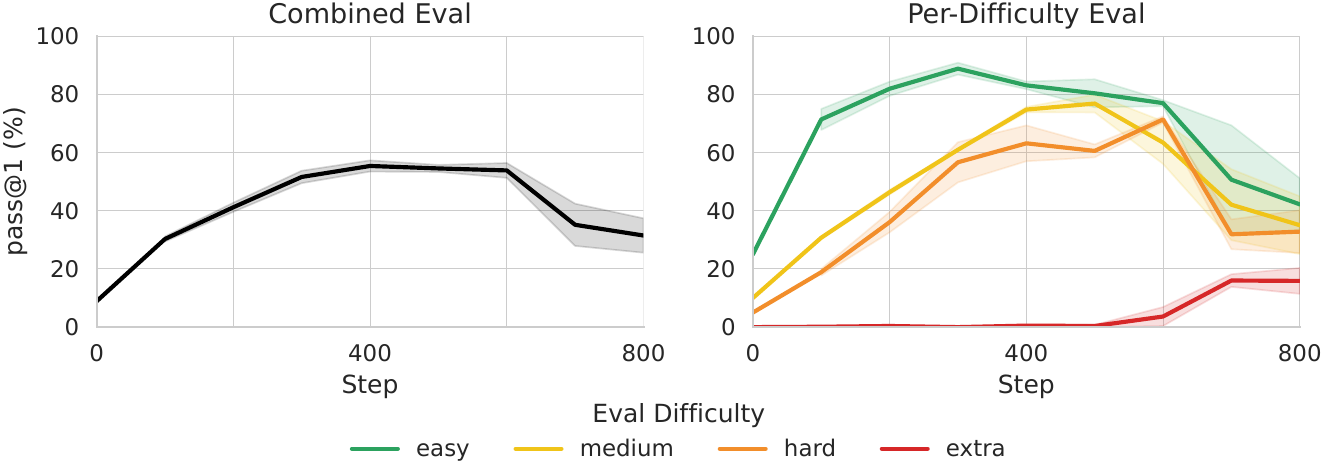}
    \vspace{-1em}
    \caption{\textbf{GSM8k with No Resampling of Easy Prompts.} Following \citet{an_polaris_2025}, we don't resample any prompts that achieve a perfect 16/16 correct completions during training. This fundamentally shifts the data distribution and we quickly see degradation on easy problems as our model shifts to answering exclusively harder prompts.}
    \label{fig:gsm8k_ngu_nores_passat1}
\end{figure}

In \autoref{fig:gsm8k_ngu_nores_passat1}, we run our baseline but shift our prompt distribution by not resampling any prompt that we solve in $\nicefrac{16}{16}$ completions, following \citet{an_polaris_2025}. Even early in training we see that our easiest subset of prompts starts to degrade in performance and eventually all subsets are degrading other than the hardest subset (red). Our takeaway is that not resampling easy prompts can be detrimental to their performance over multiple epochs and we prefer to avoid explicitly and irreversibly changing our prompt distribution.

\subsection{Deepscaler}
\label{app:extra_deepscaler}

\paragraph{Table Results}
In \autoref{tab:deepscaler_by_dataset} we note the final pass@1 results for all methods on our eval dataset, AIME and BRUMO. Our GRPO baselines' performance is better than GRPO baselines from previous work \citep{li_jointly_2025}. Though we did not include it in the original paper, we note that we also ran $N=16,K=8$ and it underperforms all other baseline settings as it simply doesn't sample enough correct solutions on such a difficult dataset. This demonstrates how $K$ that is too small can also be detrimental to performance. In \autoref{tab:deepscaler_improvement_pass_at_1}, we show the exact values for \autoref{fig:deepscaler_ngu} i.e. improvement in pass@1 by difficulty subset.


\begin{table}[h!]
\caption{\textbf{Deepscaler final pass@1 by dataset.} We show the pass@1 over three seeds for each setting of our GRPO baselines $K$ and our NGU settings $p_{NGU}$. Results are mean $\pm$ std dev over three seeds. *Darling results are taken from the original paper which trained on a different 10k random subset.}
\label{tab:deepscaler_by_dataset}
\centering
  \begin{tabular}{lccc}
  \toprule
   & AIME 25 & BRUMO 25 & Average \\
  \midrule
  Qwen 3 4B Base & 7.63 & 16.3 & 12.6 \\
  N=2, K=64 & 20.3 $\pm$ 0.9 & 28.8 $\pm$ 0.7 & 24.5 $\pm$ 0.5 \\
  N=4, K=32 & 21.4 $\pm$ 1.3 & 29.2 $\pm$ 0.7 & 25.3 $\pm$ 0.5 \\
  N=8, K=16 & 20.0 $\pm$ 1.5 & 29.6 $\pm$ 1.4 & 24.8 $\pm$ 1.0 \\
  + NGU p=0.875 & \textbf{21.8 $\pm$ 0.2} & \textbf{31.2 $\pm$ 1.3} & \textbf{26.5 $\pm$ 0.6} \\
  + NGU p=0.75 & 21.3 $\pm$ 2.1 & \textbf{31.6 $\pm$ 1.5} & \textbf{26.4 $\pm$ 0.7} \\
  + NGU p=0.5 & \textbf{22.0 $\pm$ 0.7} & 30.1 $\pm$ 1.6 & \textbf{26.1 $\pm$ 0.6} \\
\midrule
Darling \citep{li_jointly_2025}* & 20.1 & \textbf{31.7} & 25.9 \\
  \bottomrule
  \end{tabular}
\end{table}

\begin{table}[]
    \centering
        \caption{\textbf{Deepscaler improvement in pass@1 by difficulty subset.} We plot the improvement in pass@1 over the initial Qwen 3 4B-base model after RL training, by difficulty splits. Results are mean $\pm$ std dev over three seeds. }
  \label{tab:deepscaler_improvement_pass_at_1}
  \begin{tabular}{lcccc}
  \toprule
   & Easy & Medium & Hard & Total \\
  \midrule
  N=2, K=64 & 23.3 $\pm$ 1.9 & 15.3 $\pm$ 1.5 & 2.5 $\pm$ 0.2 & 11.5 $\pm$ 0.5 \\
  N=4, K=32 & 26.0 $\pm$ 2.5 & 15.4 $\pm$ 1.8 & 2.5 $\pm$ 1.1 & 12.2 $\pm$ 0.5 \\
  N=8, K=16 & 26.8 $\pm$ 2.6 & 14.5 $\pm$ 3.6 & 1.6 $\pm$ 0.3 & 11.7 $\pm$ 1.0 \\
  + NGU p=0.875 & 26.8 $\pm$ 1.7 & 16.3 $\pm$ 0.9 & \textbf{4.3 $\pm$ 1.2} & \textbf{13.5 $\pm$ 0.6} \\
  + NGU p=0.75 & \textbf{28.1 $\pm$ 0.2} & \textbf{16.8 $\pm$ 3.3} & 3.0 $\pm$ 1.2 & \textbf{13.4 $\pm$ 0.7} \\
  + NGU p=0.5 & 25.1 $\pm$ 2.3 & \textbf{17.9 $\pm$ 2.2} & 3.3 $\pm$ 1.9 & 13.0 $\pm$ 0.6 \\
  \midrule
  + Sampling Curriculum & 22.7 $\pm$ 1.6 & 15.5 $\pm$ 1.4 & \textbf{4.0 $\pm$ 1.1} & 12.1 $\pm$ 0.8 \\ 
  \bottomrule
  \end{tabular}
\end{table}


\paragraph{Prompt-Difficulty Curriculum Baseline}
We also compare NGU to a simple baseline of curriculum learning, inspired by GVM-RAFT \citep{yao_optimizing_2025} and Reinforce-Ada-Est \citep{xiong_reinforce-ada_2025}. We sample 32 completions per prompt with our initial model on our whole training dataset to get an estimate of each prompt's difficulty. We then assign each prompt a number of completions $K$ proportional to its expected pass-rate i.e. prompts with $\frac{8}{32}$ correct, or more, are assigned $K=4$, $\frac{4}{32}$ are $K=8$, $\frac{2}{32}$ are $K=16$, and all less are $K=32$. We sample exactly this $K$ for each prompt and report results in \autoref{tab:deepscaler_improvement_pass_at_1}. This difficulty curriculum greatly improves on the hardest subset, but it comes at the cost of large degradation on the easiest subset. This is in line with \autoref{sec:ngu} experiments on the other curriculum learning method \citep{an_polaris_2025}. Overall, it slightly underperforms the GRPO baseline and implies that easy prompts may change in difficulty during training. We believe that RL should still sample a minimum amount per prompt, to maintain performance on easy problems.

\begin{figure}[h!]
    \centering
    \includegraphics[width=\linewidth]{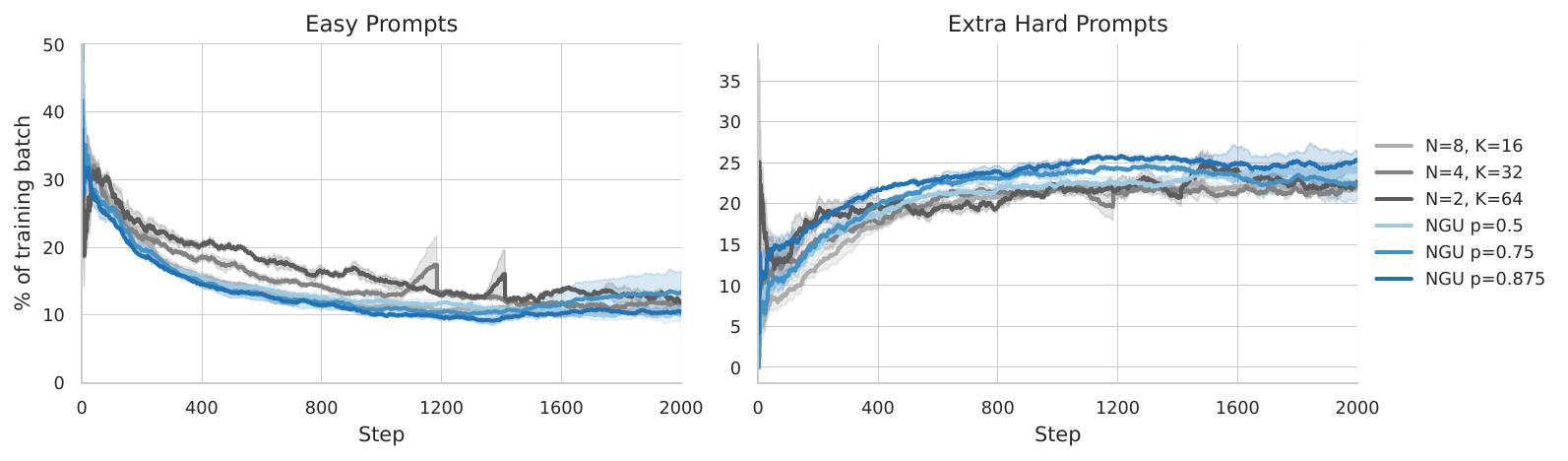}
    \caption{\textbf{NGU reallocates compute to more difficult prompts in Deepscaler.} As in GSM8k, \autoref{fig:gsm8k_baseline_nonzero}, we plot the ratios of the easiest and hardest prompts in our update batch over training. NGU increases the proportion of hard prompts and decreases the proportion of easy prompts. As before, NGU achieves the best of large $K$ early in training and small $K$ late in training.}
    \label{fig:deepscaler_prompt_ratios}
\end{figure}

\paragraph{Prompt Ratios}
In \autoref{fig:deepscaler_prompt_ratios}, we plot the composition of our training batch. We divide training prompts into four equal groups of difficulty and plot percentage of our training batch, similar to \autoref{fig:gsm8k_baseline_nonzero}. As in GSM8k, we see that NGU shifts the difficulty of prompts that we train on by having the fewest easy prompts with non-zero gradient and the most hard prompts with non-zero gradient.

\FloatBarrier
\section{Experimental Details}

For all experiments we use an asynchronous RL framework that leverages in-flight updates \citep{piche_pipelinerl_2025} to stay as on-policy as possible. We leverage Deepspeed \citep{rasley_deepspeed_2020} for training and vllm \citep{kwon_efficient_2023} for inference.    

\subsection{GSM8k}
\label{app:gsm8k}

We run GSM8k for 2000 steps using 4xL40s GPUs which takes approximately 10 hours. We note all important hyperparameters in \autoref{tab:qwen25-05b-gsm8k-hparams}

\begin{table}[h!]
\centering
\small
\caption{\textbf{Hyperparameters for the Qwen2.5-0.5B-Instruct on GSM8K} }
\label{tab:qwen25-05b-gsm8k-hparams}
\begin{tabular}{ll}
\hline
\textbf{Hyperparameter} & \textbf{Value} \\
\hline
KL beta & 0.0 \\
Learning rate & $1 \times 10^{-6}$ \\
Total Steps & 2000 \\
Temperature & 1.0 \\
Completions Per Prompt $K$ & 16 \\
Prompts per Batch $N$  & 32 \\
Advantage normalization & centered \\
Async steps & 1 \\
Prompt length & 2048 \\
Response length & 4096 \\
Clip higher & 0.28 \\
Non-stop penalty & False \\
Mask truncated completions & False \\
\hline
\end{tabular}
\end{table}

\subsection{Deepscaler}
\label{app:deepscaler}

We reproduce the Math RLVR setup of \citep{li_jointly_2025} and achieve better baseline results. We take a random 10k subset of the Deepscaler dataset \citep{luo_deepscaler_2025}. We check overlap of problems with our evals and find them to be minimal. 

\begin{figure}
    \centering
    \includegraphics[width=0.8\linewidth]{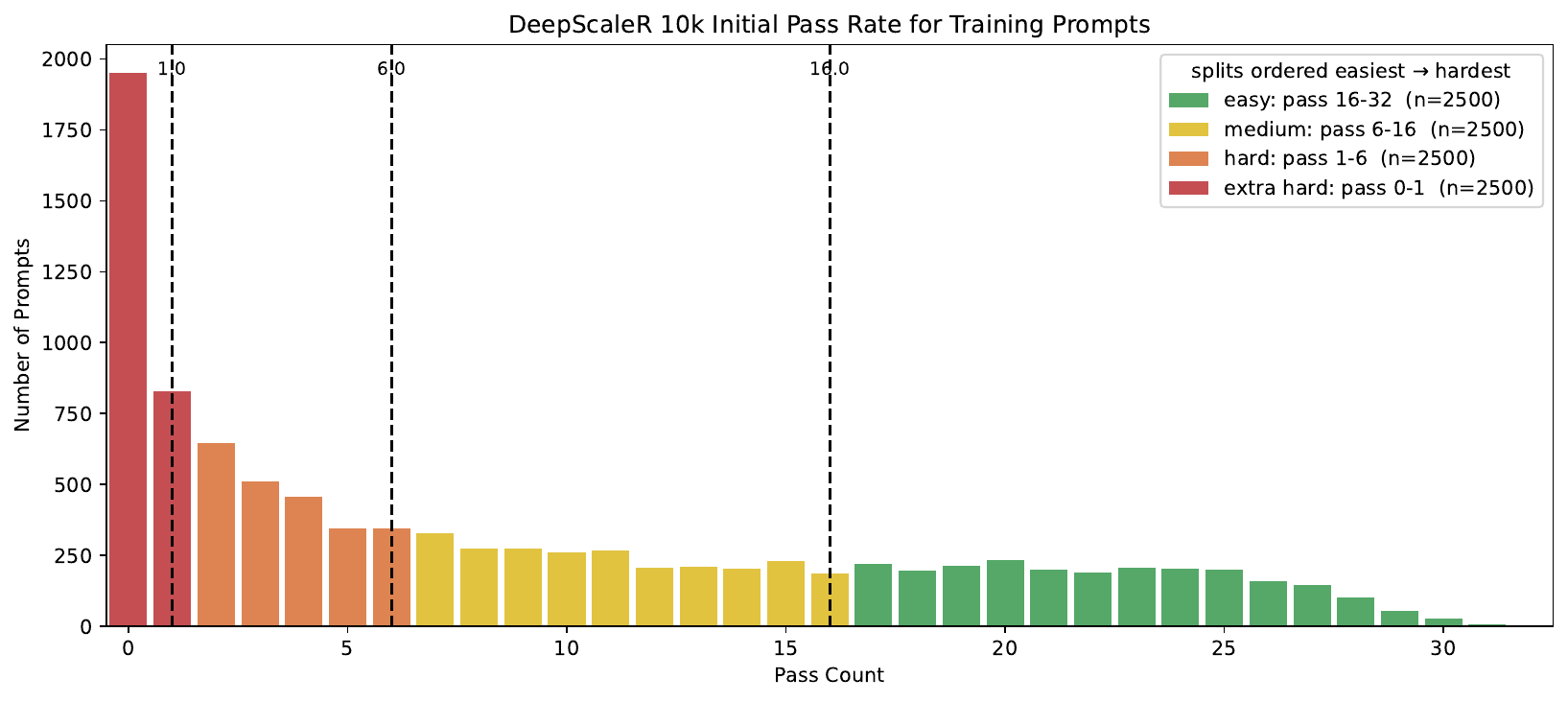}
    \caption{Difficulty of Deepscaler 10k dataset for Qwen 3 4B base. We take 32 samples on each prompt in the dataset and calculate the overall difficulty of the dataset. We find that a significant amount of data has pass@32 = 0 for the initial model. Dashed lines indicate borders between difficulty levels.}
    \label{fig:deepscaler_difficulty}
\end{figure}

To evaluate the difficulty of our prompts and evals, we run the initial model using the Qwen 3 recommended temperature 0.7 and top-p 0.8 \citep{yang_qwen3_2025}. When actually evaluating our RL-trained model, we find that the optimal temperature and top p are both simply 1. Our fastest run of 1000 steps uses 8xH100 GPUs for approximately 15 hours. We note all hyperparameters in \autoref{tab:qwen3-4b-dapo-math-hparams}. We show the difficulty of our Deepscaler 10k training dataset in \autoref{fig:deepscaler_difficulty}, including how the difficulty splits used for \autoref{fig:deepscaler_prompt_ratios}.

\begin{table}[h!]
\centering
\small
\caption{\textbf{Hyperparameters for Qwen3-4B-Base on Deepscaler Math}}
\label{tab:qwen3-4b-dapo-math-hparams}
\begin{tabular}{ll}
\hline
\textbf{Hyperparameter} & \textbf{Value} \\
\hline
KL beta & 0.0 \\
Learning rate & $1 \times 10^{-6}$ \\
Total Steps & 1000 \\
Temperature & 1.0 \\
Completions Per Prompt & 16 \\
Prompts per Batch & 8 \\
Advantage normalization & centered \\
Async steps & 4 \\
Prompt length & 2048 \\
Response length & 8192 \\
Clip higher & 0.272 \\
Non-stop penalty & False \\
Mask truncated completions & False \\
\hline
\end{tabular}
\end{table}

\FloatBarrier
\subsection{Manufactoria}
\label{app:manufactoria}

We follow the original setup of \citet{sun_rl_2025}. We note that the true setup is actually slightly different from the details published in the Manufactoria paper and code, as discovered in communication with the original authors. We note all hyperparameters in \autoref{tab:qwen3-4b-manufactoria-hparams}. Each run of 3000 steps uses 2 nodes of 8xH100 GPUs for approximately 28 hours. 

\begin{table}[h!]
\centering
\small
\caption{\textbf{Hyperparameters for Qwen3-4B-Instruct-2507 on Manufactoria}}
\label{tab:qwen3-4b-manufactoria-hparams}
\begin{tabular}{ll}
\hline
\textbf{Hyperparameter} & \textbf{Value} \\
\hline
KL beta & 0.01 \\
Learning rate & $5 \times 10^{-7}$ \\
Total Steps & 1500 \\
Temperature & 1.0 \\
Completions Per Prompt & 16 \\
Prompts per Batch & 32 \\
Advantage normalization & centered \\
Async steps & 1 \\
Prompt length & 2048 \\
Response length & 12000 \\
Clip higher & 0.28 \\
Non-stop penalty & False \\
Mask truncated completions & False \\
\hline
\end{tabular}
\end{table}

\FloatBarrier
\newpage
\section{Pseudocode}
\label{app:pseudocode}

\begin{algorithm}[h!]
\caption{\textsc{Never Give Up} Sampling for Asynchronous RL Trainer}
\label{alg:ngu}
\begin{algorithmic}[1]
\Require Generator $\mathcal{G}$, Trainer $\mathcal{L}$, completions per prompt $K$ 
\Require continuation probability $p_{\mathrm{NGU}}$, previous sample buffer $\mathcal{S}$, NGU age cutoff $T$
\While{trainer thread is running}
    \State Receive completions $y_{1:K} \sim \pi_\theta(\cdot \mid x)$ from generator $\mathcal{G}$
    \State Compute rewards $r = \{r_i \gets R(x,y_i)$ for $i=1,\ldots,K \}$
    \State Compute average reward $\bar r = \frac{1}{K} \sum_{i \in 1...K} r_i$
    \State Init count $k = K$
    \If{$\exists$ previous completions for this prompt $y_\text{NGU} \in S(x)$}
        \State Pop previous completions $y_\text{NGU}$, rewards $r_\text{NGU}$, count $k_\text{NGU}$, baseline $ \bar r_{\text{NGU}} \gets \mathcal{S}(x)$
        \State Update baseline $\bar r \gets \frac{\bar r_{\text{NGU}} * k_\text{NGU} + \bar r * K}{k_\text{NGU} + K} $
        \State Update count $k \mathrel{+}= k_{\text{NGU}}$
        \State Extend with non-stale completions $y \mathrel{\|}= y_i \in y_\text{NGU}, \text{age}(y_i) \leq T$
        \State Extend with non-stale rewards $r \mathrel{\|}= r_i \in r_\text{NGU}, \text{age}(y_i) \leq T$  
    \EndIf
    \State
    \If{$r_i = 1\ \forall_{r_i \in R}$} \Comment{All completions correct; prompt is too easy}
        \State \textbf{Give Up} on $x$
        \State 
    \ElsIf{$\exists_{i \in [1,K]} \ s.t.\ r_i > \bar r$} \Comment{Mixed outcomes provide GRPO gradient}
        \State Pass $(x, y, r, \bar r)$ for GRPO update
        \State
    \Else{} \Comment{All completions wrong; prompt may be hard}
        \State Sample $u \sim \mathrm{Uniform}(0,1)$
        \If{$u < p_{\mathrm{NGU}}$} \Comment{Never Give Up}
            \State Add prompt $x$ back to queue of generator $\mathcal{G}$
            \State Add updated completions, baseline, count to buffer $S(x) \gets y, k, \bar r$
        \Else
            \State \textbf{Give Up} on $x$
        \EndIf
    \EndIf
\EndWhile
\end{algorithmic}
\end{algorithm}


\end{document}